\documentclass[10pt,twocolumn,letterpaper]{article}

\usepackage{cvpr}
\usepackage{times}
\usepackage{epsfig}
\usepackage{graphicx}
\usepackage{amsmath}
\usepackage{amssymb}
\usepackage{booktabs}
\usepackage{comment}
\usepackage{subfigure}
\usepackage[normalem]{ulem}
\usepackage{multirow}

\usepackage{listings}
\usepackage{color}

\usepackage{mathtools}

\usepackage[pagebackref=true,breaklinks=true,letterpaper=true,colorlinks,bookmarks=false]{hyperref}

\definecolor{dkgreen}{rgb}{0,0.6,0}
\definecolor{gray}{rgb}{0.5,0.5,0.5}
\definecolor{mauve}{rgb}{0.58,0,0.82}

\usepackage[breaklinks=true,bookmarks=false]{hyperref}

\cvprfinalcopy 

\begin{document}

\title{Searching Central Difference Convolutional Networks for Face Anti-Spoofing}

\author{Zitong Yu\textsuperscript{1}, Chenxu Zhao\textsuperscript{2}, Zezheng Wang\textsuperscript{3}, Yunxiao Qin\textsuperscript{4}, \\ Zhuo Su\textsuperscript{1}, Xiaobai Li\textsuperscript{1}, Feng Zhou\textsuperscript{3}, Guoying Zhao\textsuperscript{1\thanks{denotes corresponding author}}\\
\normalsize{\textsuperscript{1}CMVS, University of Oulu  \qquad  \textsuperscript{2}Mininglamp Academy of Sciences, Mininglamp Technology}\\
\normalsize{\textsuperscript{3}Aibee \qquad
\textsuperscript{4}Northwestern Polytechnical University}\\
\tt\small \{zitong.yu, zhuo.su, xiaobai.li, guoying.zhao\}@oulu.fi, \\
\tt\small\{zhaochenxu\}@mininglamp.com,\{zezhengwang, fzhoug\}@aibee.com, \{qyxqyx\}@mail.nwpu.edu.cn
}


\maketitle

\begin{abstract}

Face anti-spoofing (FAS) plays a vital role in face recognition systems. Most state-of-the-art FAS methods 1) rely on stacked convolutions and expert-designed network, which is weak in describing detailed fine-grained information and easily being ineffective when the environment varies (e.g., different illumination), and 2) prefer to use long sequence as input to extract dynamic features, making them difficult to deploy into scenarios which need quick response. Here we propose a novel frame level FAS method based on Central Difference Convolution (CDC), which is able to capture intrinsic detailed patterns via aggregating both intensity and gradient information. A network built with CDC, called the Central Difference Convolutional Network (CDCN), is able to provide more robust modeling capacity than its counterpart built with vanilla convolution. Furthermore, over a specifically designed CDC search space, Neural Architecture Search (NAS) is utilized to discover a more powerful network structure (CDCN++), which can be assembled with Multiscale Attention Fusion Module (MAFM) for further boosting performance. Comprehensive experiments are performed on six benchmark datasets to show that 1) the proposed method not only achieves superior performance on intra-dataset testing (especially 0.2\% ACER in Protocol-1 of OULU-NPU dataset), 2) it also generalizes well on cross-dataset testing (particularly 6.5\% HTER from CASIA-MFSD to Replay-Attack datasets). The codes are available at  \href{https://github.com/ZitongYu/CDCN}{https://github.com/ZitongYu/CDCN}.

\end{abstract}

\vspace{-1.0em}
\section{Introduction}
\vspace{-0.3em}

\begin{figure}
\vspace{-0.6em}
\centering
\includegraphics[width=7.0cm,height=5.6cm]{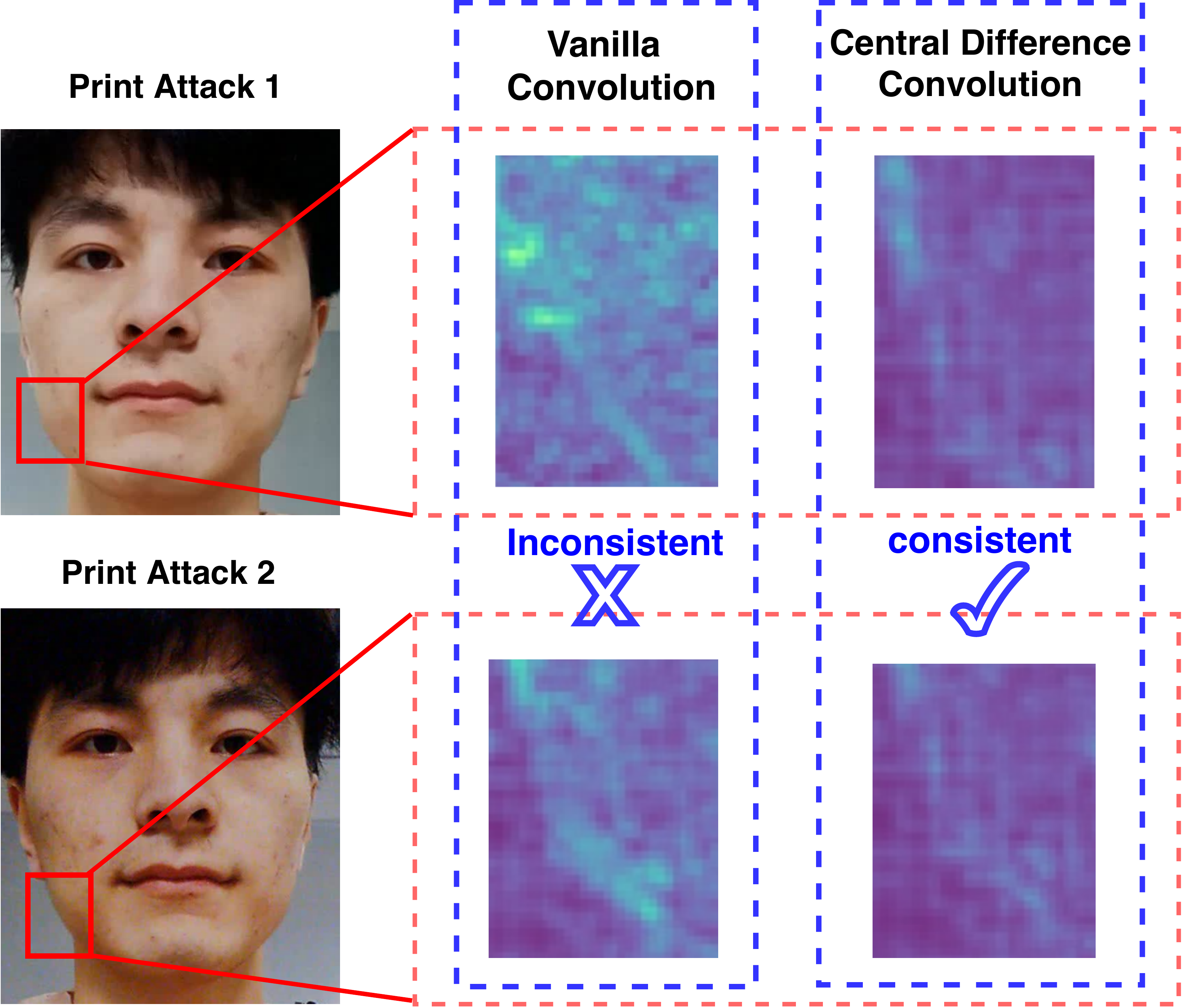}
  \caption{\small{
  Feature response of vanilla convolution (VanillaConv) and central difference convolution (CDC) for spoofing faces in shifted domains (illumination \& input camera). VanillaConv fails to capture the consistent spoofing pattern while CDC is able to extract the invariant detailed spoofing features, e.g., \textbf{lattice artifacts}.}
  }
 
\label{fig:Figure1}
\vspace{-0.6em}
\end{figure}


Face recognition has been widely used in many interactive artificial intelligence systems for its convenience. However, vulnerability to presentation attacks (PA) curtail its reliable deployment. Merely presenting printed images or videos to the biometric sensor could fool face recognition systems. Typical examples of presentation attacks are print, video replay, and 3D masks. For the reliable use of face recognition systems, face anti-spoofing (FAS) methods are important to detect such presentation attacks.


In recent years, several hand-crafted features based~\cite{boulkenafet2015face,Boulkenafet2017Face,Pereira2012LBP,Komulainen2014Context,Peixoto2011Face,Patel2016Secure} and deep learning based~\cite{qin2019learning,yu2020auto,Liu2018Learning,jourabloo2018face,yang2019face,Atoum2018Face,Gan20173D,george2019deep} methods have been proposed for presentation attack detection (PAD). On one hand, the classical hand-crafted descriptors (e.g., local binary pattern (LBP)~\cite{boulkenafet2015face}) leverage local relationship among the neighbours as the discriminative features, which is robust for describing the detailed invariant information (e.g., color texture, moir$\rm\acute{e}$ pattern and noise artifacts) between the living and spoofing faces. On the other hand, due to the stacked convolution operations with nonlinear activation, the convolutional neural networks (CNN) hold strong representation abilities to distinguish the bona fide and PA. However, CNN based methods focus on the deeper semantic features, which are weak in describing detailed fine-grained information between living and spoofing faces and easily being ineffective when the environment varies (e.g., different light illumination). \textbf{How to integrate local descriptors with convolution operation for robust feature representation is worth exploring.}

Most recent deep learning based FAS methods are usually built upon image classification task based backbones~\cite{yang2014learn,yang2019face,george2019deep}, such as VGG~\cite{simonyan2014very}, ResNet~\cite{he2016deep} and DenseNet~\cite{huang2017densely}. The networks are usually supervised by binary cross-entropy loss, which easily learns the arbitrary patterns such as screen bezel instead of the nature of spoofing patterns. In order to solve this issue, several depth supervised FAS methods ~\cite{Atoum2018Face,Liu2018Learning}, which utilize pseudo depth map label as auxiliary supervised signal, have been developed. However, all these network architectures are carefully designed by human experts, which might not be optimal for FAS task. Hence, \textbf{to automatically discover best-suited networks for FAS task with auxiliary depth supervision should be considered.}

Most existing state-of-the-art FAS methods~\cite{Liu2018Learning,wang2018exploiting,yang2019face,lin2019face} need multiple frames as input to extract dynamic spatio-temporal features (e.g., motion~\cite{Liu2018Learning,wang2018exploiting} and rPPG~\cite{yang2019face,lin2019face}) for PAD. However, long video sequence may not be suitable for specific deployment conditions where the decision needs to be made quickly. Hence, frame level PAD approaches are advantageous from the usability point of view despite inferior performance compared with video level methods. \textbf{To design high-performing frame level methods is crucial for real-world FAS applications.}

Motivated by the discussions above, we propose a novel convolution operator called Central Difference Convolution (CDC), which is good at describing fine-grained invariant information. As shown in Fig.~\ref{fig:Figure1}, CDC is more likely to extract intrinsic spoofing patterns (e.g., lattice artifacts) than vanilla convolution in diverse environments. Furthermore, over a specifically designed CDC search space, Neural Architecture Search (NAS) is utilized to discover the excellent frame level networks for depth supervised face anti-spoofing task. Our contributions include:


\begin{itemize}
\setlength\itemsep{-0.1em}
\vspace{-0.5em}
    \item We design a novel convolution operator called Central Difference Convolution (CDC), which is suitable for FAS task due to its remarkable representation ability for invariant fine-grained features in diverse environments. Without introducing any extra parameters, CDC can replace the vanilla convolution and plug and play in existing neural networks to form Central Difference Convolutional Networks (CDCN) with more robust modeling capacity.

    \item  We propose CDCN++, an extended version of CDCN, consisting of the searched backbone network and Multiscale Attention Fusion Module (MAFM) for aggregating the multi-level CDC features effectively.
    
    \item To our best knowledge, this is the first approach
    that searches neural architectures for FAS task. Different from the previous classification task based NAS supervised by softmax loss, we search the well-suited frame level networks for depth supervised FAS task over a specifically designed CDC search space.

    \item Our proposed method achieves state-of-the-art performance on all six benchmark datasets with both intra- as well as cross-dataset testing.
    
    
\end{itemize}

\vspace{-1.5em}
\section{Related Work}
\vspace{-0.5em}


\noindent\textbf{Face Anti-Spoofing.}\quad      
Traditional face anti-spoofing methods usually extract hand-crafted features from the facial images to capture the spoofing patterns. Several classical local descriptors such as LBP~\cite{boulkenafet2015face,Pereira2012LBP}, 
SIFT~\cite{Patel2016Secure}, SURF~\cite{Boulkenafet2017Face_SURF}, HOG~\cite{Komulainen2014Context} and DoG~\cite{Peixoto2011Face} are utilized to extract frame level features while video level methods usually capture dynamic clues like dynamic texture~\cite{komulainen2012face}, micro-motion~\cite{siddiqui2016face} and eye blinking~\cite{Pan2007Eyeblink}. More recently, a few deep learning based methods
are proposed for both frame level and video level face anti-spoofing. For frame level methods \cite{Li2017An,Patel2016Cross,george2019deep,jourabloo2018face}, pre-trained deep CNN models are fine-tuned to extract features in a binary-classification setting. In contrast, auxiliary depth supervised FAS methods~\cite{Atoum2018Face,Liu2018Learning} are introduced to learn more detailed information effectively. On the other hand, several video level CNN methods are presented to exploit the dynamic spatio-temporal~\cite{wang2018exploiting,yang2019face,lin2018live} or rPPG~\cite{li2016generalized,Liu2018Learning,lin2019face} features for PAD. Despite achieving state-of-the-art performance, video level deep learning based methods need long sequence as input. In addition, compared with traditional descriptors, CNN overfits easily and is hard to generalize well on unseen scenes.  


\noindent\textbf{Convolution Operators.}\quad  
The convolution operator is commonly used
in extracting basic visual features in deep learning framework. Recently extensions to the vanilla convolution operator have been proposed. In one direction, classical local descriptors (e.g., LBP \cite{ahonen2006face} and Gabor filters \cite{jain1991unsupervised}) are considered into convolution design. Representative works include Local Binary Convolution \cite{juefei2017local} and Gabor Convolution \cite{luan2018gabor}, which is proposed for saving computational cost and enhancing the resistance to the spatial changes, respectively. Another direction is to modify the spatial scope for aggregation. Two related works are dialated convolution \cite{yu2015multi} and deformable convolution \cite{dai2017deformable}. However, these convolution operators may not be suitable for FAS task because of the limited representation capacity for invariant fine-grained features.





\noindent\textbf{Neural Architecture Search.}\quad    
Our work is motivated by recent researches on NAS \cite{brock2017smash,dong2019searching,liu2018darts,pham2018efficient,zoph2016neural,zoph2018learning,xu2019pc}, while we focus on searching for a depth supervised model
with high performance instead of a binary classification model for face anti-spoofing task.
There are three main categories of existing NAS methods: 1) Reinforcement learning based \cite{zoph2016neural,zoph2018learning}, 2) Evolution algorithm based \cite{real2019regularized,real2017large}, and 3) Gradient based \cite{liu2018darts,xu2019pc,cai2019proxylessnas}. Most of NAS approaches search networks
on a small proxy task and transfer the found architecture to another large target task. For the perspective of computer vision applications, NAS has been developed for face recognition \cite{zhu2019neural}, action recognition \cite{peng2019video}, person ReID \cite{quan2019auto}, object detection \cite{ghiasi2019fpn} and segmentation \cite{zhang2019customizable} tasks.
To the best of our knowledge, no NAS based method has ever been proposed for face anti-spoofing task.

In order to overcome the above-mentioned drawbacks and fill in the blank, we search the frame level CNN over a specially designed search space with the new proposed convolution operator for depth-supervised FAS task.

\section{Methodology}
\label{sec:method}

In this section, we will first introduce our Central Difference Convolution in Section~\ref{sec:CDC}, then introduce the \textbf{C}entral \textbf{D}ifference \textbf{C}onvolutional \textbf{N}etworks (CDCN) for face anti-spoofing in Section~\ref{sec:CDCN}, and at last present the searched networks with attention mechanism (CDCN++) in Section~\ref{sec:CDCN++}. 

\subsection{Central Difference Convolution}
\label{sec:CDC}

In modern deep learning frameworks, the feature maps and convolution are represented in 3D shape (2D spatial domain and extra channel dimension). As the convolution operation remains the same across the channel dimension, for simplicity, in this subsection the convolutions are described in 2D while extension to 3D is straightforward.

\textbf{Vanilla Convolution.}\quad   
As 2D spatial convolution is the basic operation in CNN for vision tasks, here we denote it as vanilla convolution and review it shortly first. There are two main steps in the 2D convolution: 1) \textsl{sampling} local receptive field region $\mathcal{R}$ over the input feature map $x$; 2) \textsl{aggregation} of sampled values via weighted summation. Hence, the output feature map $y$ can be formulated as

\vspace{-0.5em}
\begin{equation} 
y(p_0)=\sum_{p_n\in \mathcal{R}}w(p_n)\cdot x(p_0+p_n),
\label{eq:vanilla}
\vspace{-0.5em}
\end{equation}
where $p_0$ denotes current location on both input and output feature maps while $p_n$ enumerates the locations in $\mathcal{R}$. For instance, local receptive field region for convolution operation with 3$\times$3 kernel and dilation 1 is $\mathcal{R}=\left \{  (-1,-1),(-1,0),\cdots,(0,1),(1,1)  \right \}$.

\begin{figure}
\centering
\includegraphics[width=8.3cm,height=3.1cm]{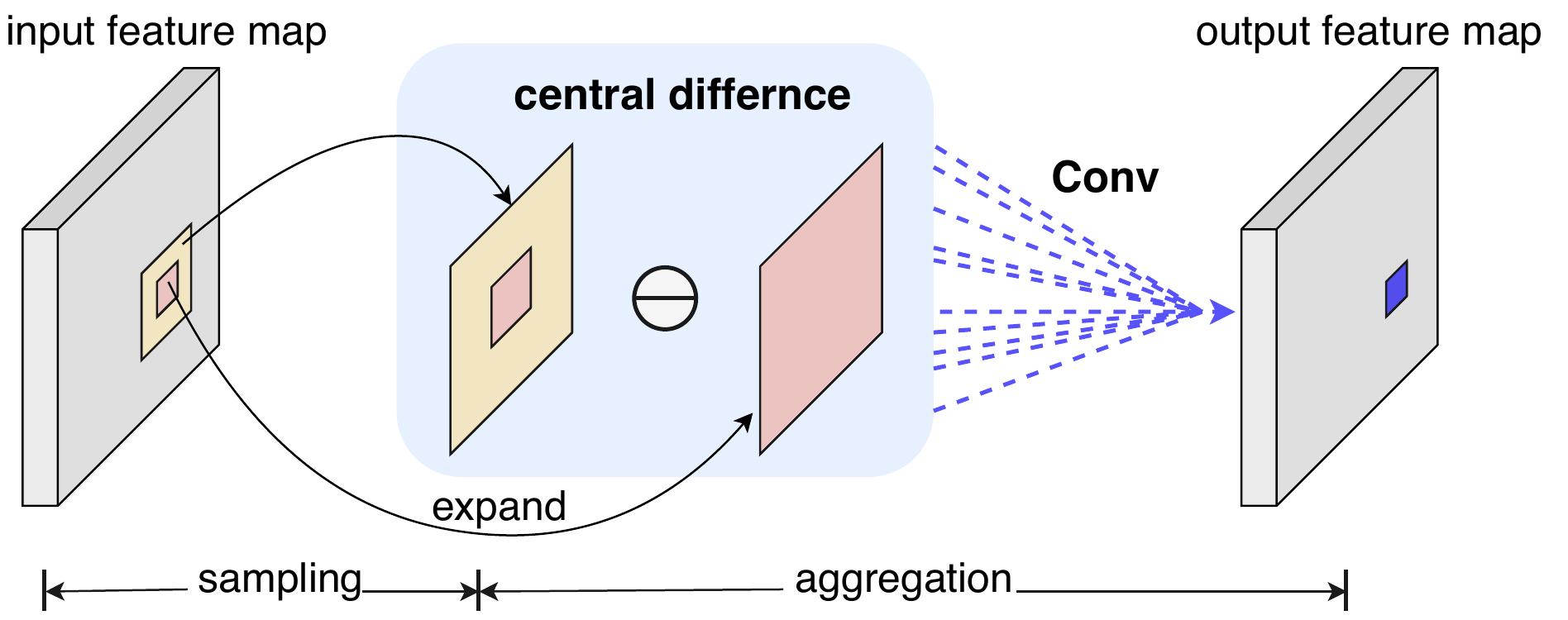}
  \caption{\small{
  Central difference convolution. 
  }
  }
 
\label{fig:CDC}
\vspace{-1.0em}
\end{figure}

\textbf{Vanilla Convolution Meets Central Difference.}\quad Inspired by the famous local binary pattern (LBP)~\cite{boulkenafet2015face} which describes local relations in a binary central difference way, we also introduce central difference into vanilla convolution to enhance its representation and generalization capacity. Similarly, central difference convolution also consists of two steps, i.e., \textsl{sampling} and \textsl{aggregation}. The sampling step is similar to that in vanilla convolution while the aggregation step is different: as illustrated in Fig.~\ref{fig:CDC}, central difference convolution prefers to aggregate the center-oriented gradient of sampled values.  Eq.~(\ref{eq:vanilla}) becomes

\vspace{-1.0em}

\begin{equation} 
y(p_0)=\sum_{p_n\in \mathcal{R}}w(p_n)\cdot (x(p_0+p_n)-x(p_0)).
\label{eq:central}
\vspace{-0.2em}
\end{equation}
When $p_n= (0,0)$, the gradient value always equals to zero with respect to the central location $p_0$ itself.

For face anti-spoofing task, both the intensity-level semantic information and gradient-level detailed message are crucial for distinguishing the living and spoofing faces, which indicates that combining vanilla convolution with central difference convolution might be a feasible manner to provide more robust modeling capacity. Therefore we generalize central difference convolution as 

\vspace{-1.0em}

\begin{equation} 
\begin{split}
y(p_0)
=\theta \cdot \underbrace{\sum_{p_n\in \mathcal{R}}w(p_n)\cdot (x(p_0+p_n)-x(p_0))}_{\text{central difference convolution}}&\\
+ (1-\theta)\cdot \underbrace{\sum_{p_n\in \mathcal{R}}w(p_n)\cdot x(p_0+p_n)}_{\text{vanilla convolution}},& \\
\end{split}
\label{eq:CDC}
\vspace{-0.2em}
\end{equation}
where hyperparameter $\theta \in [0,1]$ tradeoffs the contribution between intensity-level and gradient-level information. The higher value of $\theta$ means the more importance of central difference gradient information. We will henceforth refer to this generalized \textbf{C}entral \textbf{D}ifference \textbf{C}onvolution as \textbf{CDC}, which should be easy to identify according to its context.

\textbf{Implementation for CDC.}\quad  In order to efficiently implement CDC in modern deep learning framework, we decompose and merge Eq.~(\ref{eq:CDC}) into the vanilla convolution with additional central difference term

\vspace{-1.2em}

\begin{equation} 
\begin{split}
y(p_0) 
&= \underbrace{\sum_{p_n\in \mathcal{R}}w(p_n)\cdot x(p_0+p_n)}_{\text{vanilla convolution}}+\theta\cdot (\underbrace{-x(p_0)\cdot\sum_{p_n\in \mathcal{R}}w(p_n))}_{\text{central difference term}}.\\
\end{split}
\label{eq:CDCimplement}
\vspace{-0.6em}
\end{equation}
According to the Eq.~(\ref{eq:CDCimplement}), CDC can be easily implemented by a few lines of code in PyTorch\cite{paszke2017automatic} and TensorFlow\cite{abadi2016tensorflow}. The derivation of Eq.~(\ref{eq:CDCimplement}) and codes based on Pytorch are shown in \textsl{\textbf{Appendix A}}.

\textbf{Relation to Prior Work.}\quad  Here we discuss the relations between CDC and vanilla convolution, local binary convolution\cite{juefei2017local} and gabor convolution\cite{luan2018gabor}, which share
similar design philosophy but with different focuses. The ablation study is in Section~\ref{sec:Ablation} to show superior performance of CDC for face anti-spoofing task.


\textsl{Relation to Vanilla Convolution.} CDC is more generalized. It can be seen from Eq.~(\ref{eq:CDC}) that vanilla convolution is a special case of CDC when $\theta=0$, i.e., aggregating local intensity information without gradient message.  

\textsl{Relation to Local Binary Convolution}\cite{juefei2017local}. Local binary convolution (LBConv) focuses on computational reduction so its modeling capacity is limited. CDC focuses on enhancing rich detailed feature representation capacity without any additional parameters. On the other side, LBConv uses pre-defined filters to describe the local feature relation while CDC can learn these filters automatically.

\textsl{Relation to Gabor Convolution}\cite{luan2018gabor}. Gabor convolution (GaborConv) devotes to enhancing the representation capacity of spatial transformations (i.e.,  orientation and scale changes) while CDC focuses more on  representing fine-grained robust features in diverse environments.  


\subsection{CDCN}
\label{sec:CDCN}

Depth-supervised face anti-spoofing methods~\cite{Liu2018Learning,Atoum2018Face} take advantage of the discrimination between spoofing and living faces based on 3D shape, and provide pixel-wise detailed information for the FAS model to capture spoofing cues. 
Motivated by this, a similar depth-supervised network~\cite{Liu2018Learning} called ``DepthNet" is built up as baseline in this paper. In order to extract more fine-grained and robust features for estimating the facial depth map, CDC is introduced to form \textbf{C}entral \textbf{D}ifference \textbf{C}onvolutional \textbf{N}etworks (CDCN). Note that DepthNet is the special case of the proposed CDCN when $\theta=0$ for all CDC operators.

The details of CDCN are shown in Table~\ref{tab:network}.
Given a single RGB facial image with size $3 \times 256 \times 256$, multi-level (low-level, mid-level and high-level) fused features are extracted for predicting the grayscale facial depth with size $32 \times 32$. We use $\theta=0.7$ as the default setting and ablation study about $\theta$ will be shown in Section~\ref{sec:Ablation}.

For the loss function, mean square error loss $\mathcal{L}_{MSE}$ is utilized for pixel-wise supervision. Moreover, for the sake of fine-grained supervision needs in FAS task, contrastive depth loss $\mathcal{L}_{CDL}$~\cite{wang2018exploiting} is considered to help the networks learn more detailed features. So the overall loss $L_{overall}$ can be formulated as $\mathcal{L}_{overall}=\mathcal{L}_{MSE}+\mathcal{L}_{CDL}$.

\begin{table}[t]\small
    \centering
    
    \resizebox{0.43\textwidth}{!}{
        \begin{tabular}{c|c|c|c}
			\hline
			Level & Output & DepthNet~\cite{Liu2018Learning} & CDCN ($\theta=0.7$)  \\
			\hline 
			 & $256\times 256$ & $3\times 3 \textrm{ conv}, 64$ & $3\times 3 \textbf{ CDC}, 64$  \\
			 
			 \hline 
			 Low & $128\times 128$ & $\begin{bmatrix*}[l]
                3\times 3 \textrm{ conv}, 128\\ 
                3\times 3 \textrm{ conv}, 196\\ 
                3\times 3 \textrm{ conv}, 128\\ 
                3\times 3 \textrm{ max pool}
                \end{bmatrix*}$
                & $\begin{bmatrix*}[l]
                3\times 3\textbf{ CDC}, 128\\ 
                3\times 3\textbf{ CDC}, 196\\ 
                3\times 3\textbf{ CDC}, 128\\ 
                3\times 3\textrm{ max pool}
                \end{bmatrix*}$   \\
			 
			 \hline 
			 Mid & $64\times 64$ &$\begin{bmatrix*}[l]
                3\times 3 \textrm{ conv}, 128\\ 
                3\times 3 \textrm{ conv}, 196\\ 
                3\times 3 \textrm{ conv}, 128\\ 
                3\times 3 \textrm{ max pool}
                \end{bmatrix*}$ 
                & $\begin{bmatrix*}[l]
                3\times 3 \textbf{ CDC}, 128\\ 
                3\times 3 \textbf{ CDC}, 196\\ 
                3\times 3 \textbf{ CDC}, 128\\ 
                3\times 3 \textrm{ max pool}
                \end{bmatrix*}$  \\
                
			 \hline 
			 High & $32\times 32$ & $\begin{bmatrix*}[l]
                3\times 3 \textrm{ conv}, 128\\ 
                3\times 3 \textrm{ conv}, 196\\ 
                3\times 3 \textrm{ conv}, 128\\ 
                3\times 3 \textrm{ max pool}
                \end{bmatrix*}$ 
                & $\begin{bmatrix*}[l]
                3\times 3 \textbf{ CDC}, 128\\ 
                3\times 3 \textbf{ CDC}, 196\\ 
                3\times 3 \textbf{ CDC}, 128\\ 
                3\times 3 \textrm{ max pool}
                \end{bmatrix*}$   \\
            
            \hline 
             & $32\times 32$ & \multicolumn{2}{c}{[concat (Low, Mid, High), $384$]}  \\
            
            \hline 
			 & $32\times 32$ & $\begin{bmatrix*}[l]
                3\times 3 \textrm{ conv}, 128\\ 
                3\times 3 \textrm{ conv}, 64\\ 
                3\times 3 \textrm{ conv}, 1 
                \end{bmatrix*}$ 
                & $\begin{bmatrix*}[l]
                3\times 3 \textbf{ CDC}, 128\\ 
                3\times 3 \textbf{ CDC}, 64\\ 
                3\times 3 \textbf{ CDC}, 1
                \end{bmatrix*}$   \\
            
            \hline 
            \multicolumn{2}{c|}{\# params} &  $2.25\times 10^6$ &  $2.25\times 10^6$ \\
            

			\hline
		\end{tabular}
}
\vspace{0.1em}
\caption{Architecture of DepthNet and CDCN. Inside the brackets are the filter sizes and feature dimensionalities. ``conv" and ``CDC" suggest vanilla and central difference convolution, respectively. All convolutional layers are with stride=1 and followed by a BN-ReLU layer while max pool layers are with stride=2.}	
\vspace{-0.9em}
\label{tab:network}
\end{table}


\subsection{CDCN++}
\label{sec:CDCN++}

It can be seen from Table~\ref{tab:network} that the architecture of CDCN is designed coarsely (e.g., simply repeating the same block structure for different levels), which might be sub-optimized for face anti-spoofing task. Inspired by the classical visual object understanding models \cite{palmeri2004visual}, we propose an extended version CDCN++ (see Fig.~\ref{fig:CDCNplus}), which consists of a NAS based backbone and Multiscale Attention Fusion Module (MAFM) with selective attention capacity. 


\textbf{Search Backbone for FAS task.}\quad  Our searching algorithm is based on two gradient-based NAS methods \cite{liu2018darts,xu2019pc}, and more technical details can be referred to the original papers. Here we mainly state the new contributions about searching backbone for FAS task.

\begin{figure}
\centering
\includegraphics[width=8.0cm,height=6.2cm]{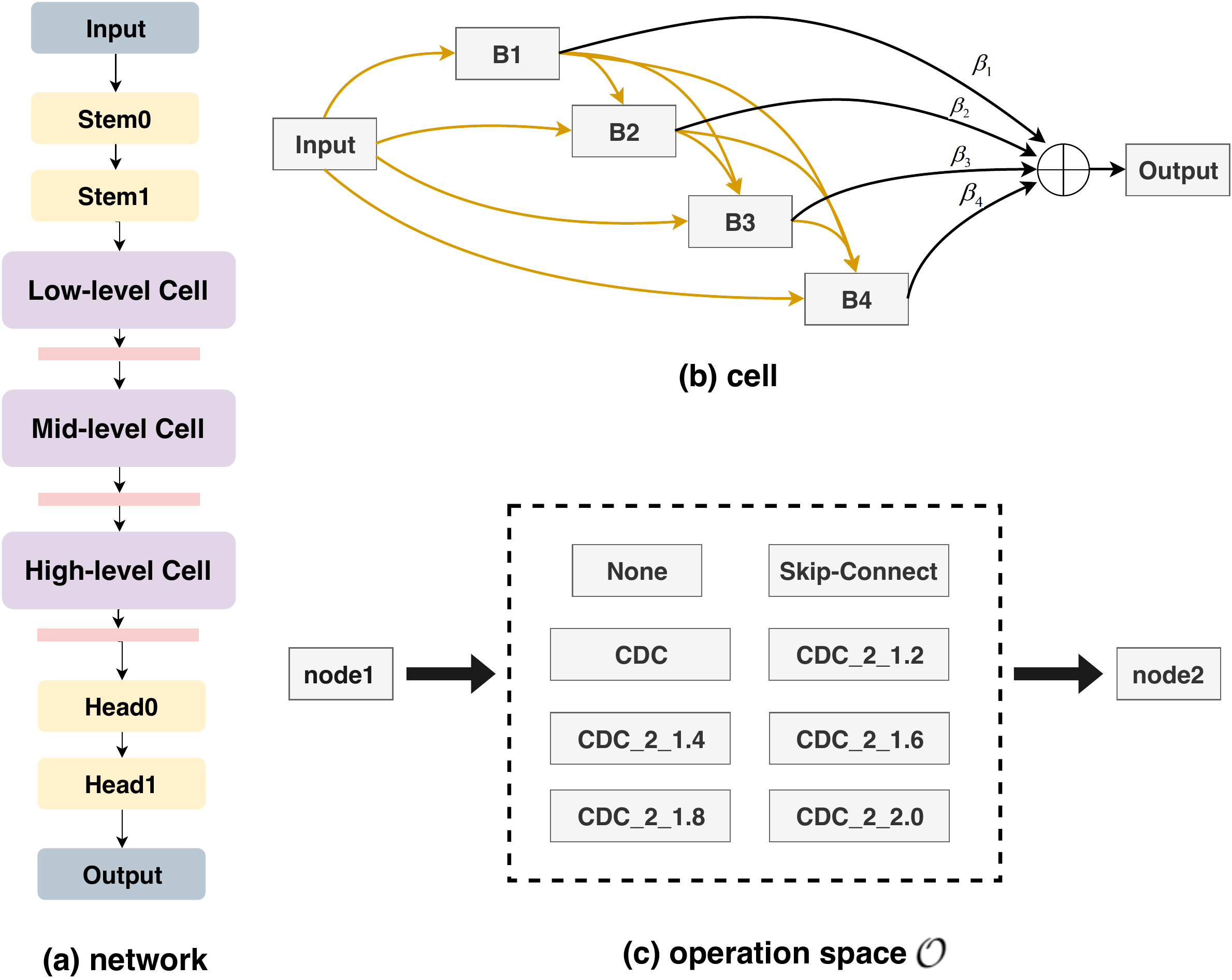}
  \caption{\small{
  Architecture search space with CDC. (a) A network consists of three stacked cells with max pool layer while stem and head layers adopt CDC with 3$\times$3 kernel and $\theta=0.7$. (b) A cell contains 6 nodes, including an input node, four intermediate nodes B1, B2, B3, B4 and an output node. (c) The edge between two nodes (except output node) denotes a possible operation. The operation space consists of eight candidates, where \textbf{CDC\_2\_$r$} means using two stacked CDC to increase channel number with ratio $r$ first and then decrease back to the original channel size. The size of total search space is $3\times8^{(1+2+3+4)}=3\times8^{10}$.}
  }
 
\label{fig:searchspace}
\vspace{-0.5em}
\end{figure}

As illustrated in Fig.~\ref{fig:searchspace}(a), the goal is to search for cells in three levels (low-level, mid-level and high-level) to form a network backbone for FAS task. Inspired by the dedicated neurons for hierarchical organization in human visual system \cite{palmeri2004visual}, we prefer to search these multi-level cells freely (i.e., cells with varied structures), which is more flexible and generalized. We name this configuration as ``\textbf{Varied Cells}" and will study its impacts in Sec. ~\ref{sec:Ablation} (see Tab.~\ref{tab:NAS}). Different from previous works \cite{liu2018darts,xu2019pc}, we adopt only one output of the latest incoming cell as the input of the current cell.


As for the cell-level structure, Fig.~\ref{fig:searchspace}(b) shows that each cell is represented as a directed acyclic graph (DAG) of $N$ nodes $\left \{ x \right \}^{N-1}_{i=0}$, where each node represents a network layer. We denote the operation space as $\mathcal{O}$, and Fig.~\ref{fig:searchspace}(c) shows eight designed candidate operations (none, skip-connect and CDCs). Each edge $(i,j)$ of DAG represents the information flow from node $x_{i}$ to node $x_{j}$, which consists of the candidate operations weighted by the architecture parameter $\alpha^{(i,j)}$. Specially, each edge $(i,j)$ can be formulated by a function $\tilde{o}^{(i,j)}$ where $\tilde{o}^{(i,j)}(x_i)=\sum_{o\in \mathcal{O}}\eta_{o}^{(i,j)}\cdot o(x_{i})$. Softmax function is utilized to relax architecture parameter $\alpha^{(i,j)}$ into operation weight $o\in \mathcal{O}$, that is $\eta_{o}^{(i,j)}=\frac{exp(\alpha_{o}^{(i,j)})}{\sum_{{o}'\in \mathcal{O}}exp(\alpha_{{o}'}^{(i,j)})}$. The intermediate node can be denoted as $x_{j}=\sum_{i<j}{\tilde{o}}^{(i,j)}(x_{i})$ and the output node $x_{N-1}$ is represented by weighted summation of all intermediate nodes $x_{N-1}=\sum_{0<i<N-1}\beta_{i}(x_{i})$. Here we propose a node attention strategy to learn the importance weights $\beta$ among intermediate nodes, that is $\beta_{i}=\frac{exp(\beta'_{i} )}{\sum_{0<j<N-1}exp(\beta'_{j})}$, where $\beta_{i}$ is the softmax of the original learnable weight ${\beta_{i}}'$ for intermediate node $x_{i}$. 


\vspace{0.2em}

In the searching stage, $\mathcal{L}_{train}$ and $\mathcal{L}_{val}$ are denoted as
the training and validation loss respectively, which are all based on the depth-supervised loss $\mathcal{L}_{overall}$ described in Sec. \ref{sec:CDCN}. Network parameters $w$ and architecture parameters $\alpha$ are learned with the following bi-level optimization problem:

\vspace{-1.5em}

\begin{equation} 
\begin{split}
&\underset{\alpha}{min} \quad \mathcal{L}_{val}(w ^{*}(\alpha),\alpha ), \\
&s.t. \quad w  ^{*}(\alpha)=arg  \,
\underset{w }{min}   \;
\mathcal{L}_{train}(w ,\alpha)
\end{split}
\label{eq:optimize}
\vspace{-2.1em}
\end{equation}

After convergence, the final discrete architecture is derived by: 1) setting $o^{(i,j)}=arg\,max_{o\in \mathcal{O},o\neq none}\,p_{o}^{(i,j)}$; 2) for each intermediate node, choosing one incoming edge with the largest value of $max_{o\in \mathcal{O},o\neq none}\,p_{o}^{(i,j)}$, and 3) for each output node, choosing the one incoming intermediate node with largest value of $max_{0<i<N-1}\,\beta_{i}$ (denoted as ``\textbf{Node Attention}") as input. In contrast, choosing last intermediate node as output node is more straightforward. We will compare these two settings in Sec. ~\ref{sec:Ablation} (see Tab.~\ref{tab:NAS}).  


\vspace{0.2em}
\textbf{MAFM.}\quad  Although simply fusing low-mid-high levels features can boost performance for the searched CDC architecture, it is still hard to find the important regions to focus, which goes against learning more discriminative features. Inspired by the selective attention in human visual system \cite{palmeri2004visual,ungerleider2000mechanisms}, neurons at different levels are likely to have stimuli in their receptive fields with various attention. Here we propose a Multiscale Attention Fusion Module (MAFM), which is able to refine and fuse low-mid-high levels CDC features via spatial attention.

\begin{figure}
\centering
\includegraphics[width=8.6cm,height=3.8cm]{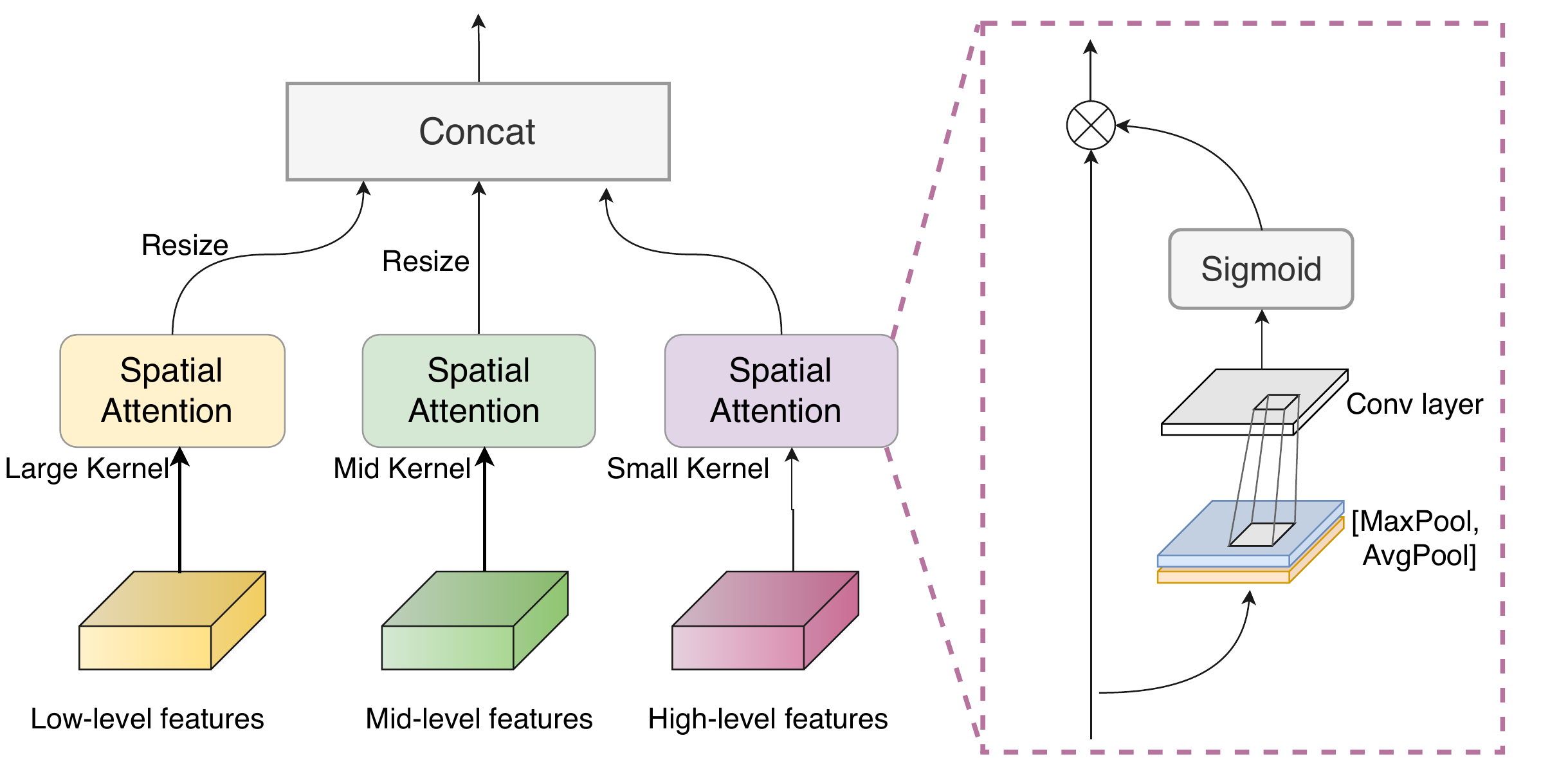}
  \caption{\small{
  Multiscale Attention Fusion Module.}
  }
 
\label{fig:Attention}
\vspace{-1.0em}
\end{figure}

As illustrated in Fig.~\ref{fig:Attention}, features $\mathcal{F}$ from different levels are refined via spatial attention \cite{woo2018cbam} with receptive fields related kernel size (i.e., the high/semantic level should be with small attention kernel size while low level with large attention kernel size in our case) and then concatenate together. The refined features $\mathcal{F}'$ can be formulated as 

\vspace{-1.2em}
\begin{equation} 
\mathcal{F}'_{i}=\mathcal{F}_{i}\, \odot \,(\sigma  (\mathcal{C}_{i}([\mathcal{A}(\mathcal{F}_{i}),\mathcal{M}(\mathcal{F}_{i})]))),i\in \left \{low,mid,high \right \},
\label{eq:attention}
\end{equation}
where $\odot$ represents Hadamard product. $\mathcal{A}$ and $\mathcal{M}$ denotes avg and max pool layer respectively. $\sigma$ means the sigmoid function while $\mathcal{C}$ is the convolution layer. Vanilla convolutions with 7$\times$7, 5$\times$5 and 3$\times$3 kernels are utilized for $\mathcal{C}_{low}$, $\mathcal{C}_{mid}$ and $\mathcal{C}_{high}$, respectively. CDC is not chosen here because of its limited capacity of global semantic cognition, which is vital in spatial attention. The corresponding ablation study is conducted in Sec. \ref{sec:Ablation}.

\begin{figure*}
\centering
\includegraphics[width=16.5cm,height=3.0cm]{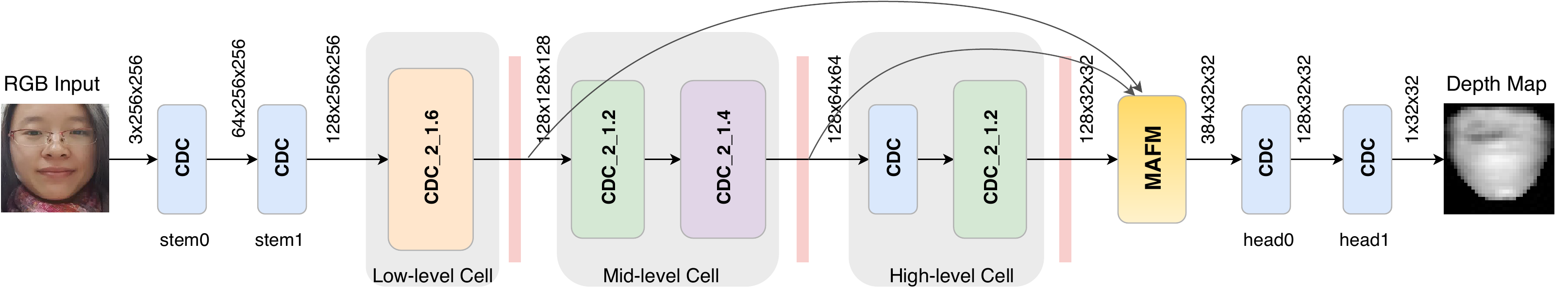}
  \caption{\small{
  The architecture of CDCN++. It consists of the searched CDC backbone and MAFM. Each cell is followed by a max pool layer.}
  }
 
\label{fig:CDCNplus}
\vspace{-1.2em}
\end{figure*}

\vspace{-0.4em}
\section{Experiments}
\vspace{-0.1em}
\label{sec:experiemnts}
In this section, extensive experiments are performed to demonstrate the effectiveness
of our method. In the following, we sequentially
describe the employed datasets \& metrics (Sec. \ref{sec:dataset}), implementation details (Sec. \ref{sec:Details}), results (Sec. \ref{sec:Ablation} - \ref{sec:Inter}) and analysis (Sec. \ref{sec:Analysis}).

\vspace{-0.1em}
\subsection{Datasets and Metrics}
\vspace{-0.1em}

\label{sec:dataset}
\textbf{Databases.} 
Six databases OULU-NPU~\cite{Boulkenafet2017OULU}, SiW~\cite{Liu2018Learning}, CASIA-MFSD~\cite{Zhang2012A}, Replay-Attack~\cite{ReplayAttack}, MSU-MFSD~\cite{wen2015face} and SiW-M~\cite{liu2019deep} are used in our experiments. OULU-NPU and SiW are high-resolution databases, containing four and three protocols to validate the generalization (e.g., unseen illumination and attack medium) of models respectively, which are utilized for intra testing. CASIA-MFSD, Replay-Attack and MSU-MFSD are databases which contain low-resolution videos, which are used for cross testing. SiW-M is designed for cross-type testing for unseen attacks as there are rich (13) attacks types inside.  

\textbf{Performance Metrics.}\quad
In OULU-NPU and SiW dataset, we follow the original protocols and metrics, i.e., Attack Presentation Classification Error Rate (APCER), Bona Fide Presentation Classification Error Rate (BPCER), and ACER~\cite{ACER} for a fair comparison. Half Total Error Rate (HTER) is adopted in the cross testing between CASIA-MFSD and Replay-Attack. Area Under Curve (AUC) is utilized for intra-database cross-type test on CASIA-MFSD, Replay-Attack and MSU-MFSD. For the cross-type test on SiW-M, APCER, BPCER, ACER and Equal Error Rate (EER) are employed.

\subsection{Implementation Details}
\vspace{-0.2em}
\label{sec:Details}

\textbf{Depth Generation.}\quad
 Dense face alignment PRNet~\cite{Feng2018Joint} is adopted to estimate the 3D shape of the living face and generate the facial depth map with size $32\times32$. More details and samples can be found in ~\cite{wang2018exploiting}. To distinguish living faces from spoofing faces, at the training stage, we normalize living depth map in a range of $[0, 1]$, while setting spoofing depth map to 0, which is similar to ~\cite{Liu2018Learning}.

\textbf{Training and Testing Setting.}\quad 
Our proposed method is implemented with Pytorch. In the training stage, models are trained with Adam optimizer and the initial learning rate (lr) and weight decay (wd) are 1e-4 and 5e-5, respectively. We train models with maximum 1300 epochs while lr halves every 500 epochs. The batch size is 56 on eight 1080Ti GPUs. In the testing stage, we calculate the mean value of the predicted depth map as the final score.   

\textbf{Searching Setting.}  Similar to \cite{xu2019pc}, partial channel connection and edge normalization are adopted. The initial number of channel is sequentially $\left \{32, 64, 128, 128, 128, 64, 1 \right \}$ in the network (see Fig.~\ref{fig:searchspace}(a)), which doubles after searching. Adam optimizer with lr=1e-4 and wd=5e-5 is utilized when training the model weights. The architecture parameters are trained with Adam optimizer with lr=6e-4 and wd=1e-3. We search 60 epochs on Protocol-1 of OULU-NPU with batchsize 12 while architecture parameters are not updated in the first 10 epochs. The whole process costs one day on three 1080Ti.


\subsection{Ablation Study}
\label{sec:Ablation}
In this subsection, all ablation studies are conducted on Protocol-1 (different illumination condition and location between train and test sets) of OULU-NPU~\cite{Boulkenafet2017OULU} to explore the details of our proposed CDC, CDCN and CDCN++.

\begin{figure}
\includegraphics[width=8.6cm,height=2.8cm]{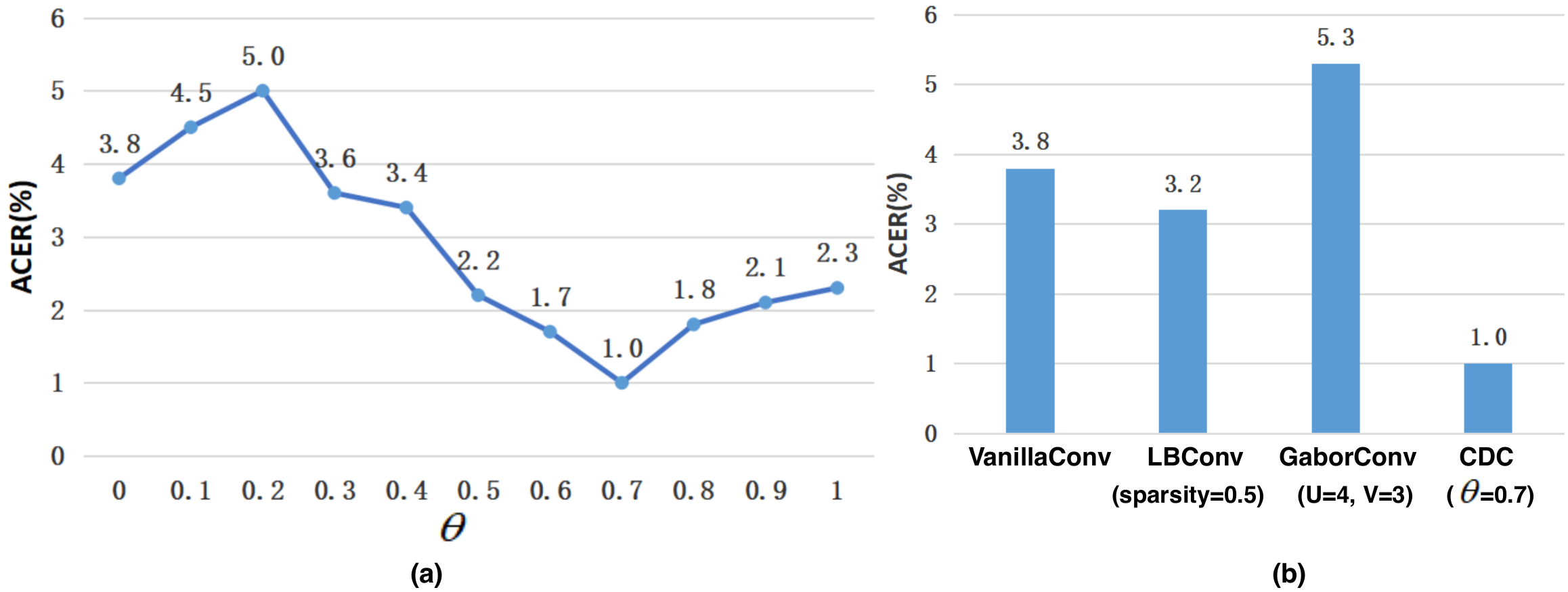}
\vspace{-1.4em}
  \caption{\small{
  (a) Impact of $\theta$ in CDCN. (b) Comparison among various convolutions (only showing the hyperparameters with best performance). The lower ACER, the better performance.}
  }
 
\label{fig:Ablation1}
\vspace{-0.4em}
\end{figure}

\textbf{Impact of $\theta$ in CDCN.}\quad 
According to Eq.~(\ref{eq:CDC}), $\theta$ controls the contribution of the gradient-based details, i.e., the higher $\theta$, the more local detailed information included. As illustrated in Fig.~\ref{fig:Ablation1}(a), when $\theta\geqslant0.3$, CDC always achieves better performance than vanilla convolution ($\theta=0$, ACER=3.8\%), indicating the central difference based fine-grained information is helpful for FAS task. As the best performance (ACER=1.0\%) is obtained when $\theta=0.7$, we use this setting for the following experiments. Besides keeping the constant $\theta$ for all layers, we also explore an adaptive CDC method to learn $\theta$ for every layer, which is shown in \textsl{\textbf{Appendix B}}.

\textbf{CDC vs. Other Convolutions.}\quad 
As discussed in Sec. \ref{sec:CDC} about the relation between CDC and prior convolutions, we argue that the proposed CDC is more suitable for FAS task as the detailed spoofing artifacts in diverse environments should be represented by the gradient-based invariant features. Fig.~\ref{fig:Ablation1}(b) shows that CDC outperforms other convolutions by a large margin (more than 2\% ACER). It is interesting to find that LBConv performs better than vanilla convolution, indicating that the local gradient information is important for FAS task. GaborConv performs the worst because it is designed for capturing spatial invariant features, which is not helpful in face anti-spoofing task.

\begin{table}[t]\small
\centering
\caption{The ablation study of NAS configuration.}
\scalebox{0.85}{\begin{tabular}{|l|c | c|c|}
\hline
Model & Varied cells & Node attention &ACER(\%)\\
\hline
NAS\_Model 1 &\   &\    & 1.7 \\
NAS\_Model 2 &$\surd$ &\   & 1.5 \\
NAS\_Model 3 & &$\surd$  & 1.4 \\
NAS\_Model 4 &$\surd$ &$\surd$  & \textbf{1.3} \\
\hline
\end{tabular}}
\label{tab:NAS}
\vspace{-1.0em}
\end{table}

\textbf{Impact of NAS Configuration.}\quad 
Table~\ref{tab:NAS} shows the ablation study about the two NAS configurations described in Sec. \ref{sec:CDCN++}, i.e., varied cells and node attention. Compared to the baseline setting with the shared cells and last intermediate node as output node, both these two configurations can boost the searching performance. The reason behind is twofold: 1) with more flexible searching constraints, NAS is able to find dedicated cells for different levels, which is more similar to human visual system \cite{palmeri2004visual}, and 2) taking the last intermediate node as output might not be optimal while choosing the most important one is more reasonable.

\begin{table}[t]
\centering
\caption{The ablation study of NAS based backbone and MAFM.}
\scalebox{0.82}{\begin{tabular}{|l|c|c|}
\hline
Backbone & Multi-level Fusion & ACER(\%)\\
\hline
w/o NAS & w/ multi-level concat & 1.0\\
w/ NAS & w/ multi-level concat & 0.7\\
\hline
w/ NAS & w/ MAFM (3x3,3x3,3x3 CDC) & 1.2\\
w/ NAS & w/ MAFM (3x3,3x3,3x3 VanillaConv) & 0.6\\
w/ NAS & w/ MAFM (5x5,5x5,5x5 VanillaConv) & 1.1\\
w/ NAS & w/ MAFM (7x7,5x5,3x3 VanillaConv) & \textbf{0.2}\\

\hline
\end{tabular}}
\label{tab:CDCD++}
\vspace{-0.8em}
\end{table}

\begin{table}[t]
\centering
\caption{The results of intra testing on four protocols of OULU-NPU. We only report the results ``STASN ~\cite{yang2019face}" trained without extra datasets for a fair comparison. }
\resizebox{0.45\textwidth}{!}{
\begin{tabular}{|c|c|c|c|c|}

\hline
Prot. & Method & APCER(\%) & BPCER(\%) & ACER(\%) \\
\hline
\multirow{6}{*}{1}
        &GRADIANT ~\cite{boulkenafet2017competition}&1.3 &12.5 & 6.9 \\
        &STASN ~\cite{yang2019face} &1.2 &2.5 & 1.9 \\
        &Auxiliary ~\cite{Liu2018Learning} &1.6 &1.6 & 1.6 \\
        &FaceDs ~\cite{jourabloo2018face} &1.2 &1.7 & 1.5 \\
        &FAS-TD ~\cite{wang2018exploiting} &2.5 &0.0 & 1.3 \\
        &DeepPixBiS ~\cite{george2019deep}&0.8 &0.0 & 0.4 \\
        &\textbf{CDCN (Ours)} &0.4 &1.7 & 1.0 \\
        &\textbf{CDCN++ (Ours)} &0.4 &0.0 & \textbf{0.2} \\
\hline
\multirow{6}{*}{2} 
       &DeepPixBiS ~\cite{george2019deep}&11.4 &0.6 & 6.0 \\
       &FaceDs ~\cite{jourabloo2018face}&4.2 &4.4 & 4.3 \\
       &Auxiliary ~\cite{Liu2018Learning}&2.7 &2.7 & 2.7 \\
       &GRADIANT ~\cite{boulkenafet2017competition}&3.1 &1.9 & 2.5 \\
       &STASN ~\cite{yang2019face}&4.2 &0.3 & 2.2 \\
       &FAS-TD ~\cite{wang2018exploiting} &1.7 &2.0 & 1.9 \\
        &\textbf{CDCN (Ours)} &1.5 &1.4 & 1.5 \\
        &\textbf{CDCN++ (Ours)} &1.8 &0.8 & \textbf{1.3} \\
\hline
\multirow{4}{*}{3} 
       &DeepPixBiS ~\cite{george2019deep}&11.7$\pm$19.6 &10.6$\pm$14.1 & 11.1$\pm$9.4 \\
       &FAS-TD ~\cite{wang2018exploiting}&5.9$\pm$1.9 &5.9$\pm$3.0 & 5.9$\pm$1.0 \\
       &GRADIANT ~\cite{boulkenafet2017competition}&2.6$\pm$3.9 &5.0$\pm$5.3 &3.8$\pm$2.4 \\
       &FaceDs ~\cite{jourabloo2018face}&4.0$\pm$1.8 &3.8$\pm$1.2 &3.6$\pm$1.6 \\
       &Auxiliary ~\cite{Liu2018Learning}&2.7$\pm$1.3 &3.1$\pm$1.7 &{2.9}$\pm$1.5 \\
       &STASN ~\cite{yang2019face}&4.7$\pm$3.9 &0.9$\pm$1.2  &2.8$\pm$1.6 \\
       &\textbf{CDCN (Ours)} &2.4$\pm$1.3 &2.2$\pm$2.0  &2.3$\pm$1.4 \\
        &\textbf{CDCN++ (Ours)} &1.7$\pm$1.5 &2.0$\pm$1.2  & \textbf{1.8$\pm$0.7} \\
\hline
\multirow{4}{*}{4} 
        &DeepPixBiS ~\cite{george2019deep}&36.7$\pm$29.7 &13.3$\pm$14.1 & 25.0$\pm$12.7 \\
       &GRADIANT ~\cite{boulkenafet2017competition}&5.0$\pm$4.5 &15.0$\pm$7.1 &10.0$\pm$5.0 \\
       &Auxiliary ~\cite{Liu2018Learning}&9.3$\pm$5.6 &10.4$\pm$6.0 &9.5$\pm$6.0 \\
       &FAS-TD ~\cite{wang2018exploiting}&14.2$\pm$8.7 &4.2$\pm$3.8 & 9.2$\pm$3.4 \\
       &STASN ~\cite{yang2019face}&6.7$\pm$10.6 &8.3$\pm$8.4  &7.5$\pm$4.7 \\
       &FaceDs ~\cite{jourabloo2018face}&1.2$\pm$6.3 &6.1$\pm$5.1 &5.6$\pm$5.7 \\
       &\textbf{CDCN (Ours)} &4.6$\pm$4.6 &9.2$\pm$8.0  &6.9$\pm$2.9 \\
       &\textbf{CDCN++ (Ours)} &4.2$\pm$3.4 &5.8$\pm$4.9  & \textbf{5.0$\pm$2.9} \\
\hline
\end{tabular}
}
\label{tab:OULU}
\vspace{-0.5em}
\end{table}

\textbf{Effectiveness of NAS Based Backbone and MAFM.}\quad 
The proposed CDCN++, consisting of NAS based backbone and MAFM, is shown in Fig.~\ref{fig:CDCNplus}. It is obvious that cells from multiple levels are quite different and the mid-level cell has deeper (four CDC) layers. Table~\ref{tab:CDCD++} shows the ablation studies about NAS based backbone and MAFM. It can be seen from the first two rows that NAS based backbone with direct multi-level fusion outperforms (0.3\% ACER) the backbone without NAS, indicating the effectiveness of our searched architecture. Meanwhile, backbone with MAFM achieves 0.5\% ACER lower than that with direct multi-level fusion, which shows the effectiveness of MAFM. We also analyse the convolution type and kernel size in MAFM and find that vanilla convolution is more suitable for capturing the semantic spatial attention. Besides, the attention kernel size should be large (7x7) and small (3x3) enough for low-level and high-level features, respectively.

\begin{table}
\centering
\caption{The results of intra testing on three protocols of SiW~\cite{Liu2018Learning}. } 
\resizebox{0.45\textwidth}{!}{
\begin{tabular}{|c|c|c|c|c|}
\hline
Prot. & Method & APCER(\%) & BPCER(\%) & ACER(\%) \\
\hline
\multirow{3}{*}{1} 
       &Auxiliary ~\cite{Liu2018Learning}&3.58 &3.58 &3.58 \\
       &STASN ~\cite{yang2019face}&-- &-- &1.00 \\
       &FAS-TD ~\cite{wang2018exploiting} &0.96 &0.50 &0.73 \\
        &\textbf{CDCN (Ours)} &0.07 &0.17 & \textbf{0.12} \\
        &\textbf{CDCN++ (Ours)}&0.07 &0.17 & \textbf{0.12} \\
\hline
\multirow{3}{*}{2} &Auxiliary ~\cite{Liu2018Learning}&0.57$\pm$0.69 &0.57$\pm$0.69 &0.57$\pm$0.69 \\
       &STASN ~\cite{yang2019face}&-- &-- &0.28$\pm$0.05 \\
       &FAS-TD ~\cite{wang2018exploiting}&0.08$\pm$0.14 &0.21$\pm$0.14 & 0.15$\pm$0.14 \\
       &\textbf{CDCN (Ours)} &0.00$\pm$0.00 &0.13$\pm$0.09  &0.06$\pm$0.04 \\
       &\textbf{CDCN++ (Ours)} &0.00$\pm$0.00 &0.09$\pm$0.10  & \textbf{0.04$\pm$0.05} \\
\hline
\multirow{3}{*}{3} &STASN ~\cite{yang2019face}&-- &-- &12.10$\pm$1.50 \\
       &Auxiliary ~\cite{Liu2018Learning}&8.31$\pm$3.81 &8.31$\pm$3.80 &8.31$\pm$3.81 \\
       &FAS-TD ~\cite{wang2018exploiting}&3.10$\pm$0.81 &3.09$\pm$0.81 & 3.10$\pm$0.81 \\
       &\textbf{CDCN (Ours)} &1.67$\pm$0.11 &1.76$\pm$0.12  &\textbf{1.71$\pm$0.11} \\
       &\textbf{CDCN++ (Ours)} &1.97$\pm$0.33 &1.77$\pm$0.10  & 1.90$\pm$0.15 \\
\hline
\end{tabular}
}
\label{tab:SiW}
\vspace{-1.0em}
\end{table}

 \vspace{-0.4em}
\subsection{Intra Testing}
 \vspace{-0.3em}
 The intra testing is carried out on both the OULU-NPU
and the SiW datasets. We strictly follow the four protocols on OULU-NPU and three protocols on SiW for
the evaluation. All compared methods including STASN ~\cite{yang2019face} are trained without extra datasets for a fair comparison. 

\textbf{Results on OULU-NPU.} \quad  As shown in Table~\ref{tab:OULU}, our proposed CDCN++ ranks first on all 4 protocols (0.2\%, 1.3\%, 1.8\% and 5.0\% ACER, respectively), which indicates the proposed method performs well at the generalization of the external environment, attack mediums and input camera variation. Unlike other state-of-the-art methods (Auxiliary~\cite{Liu2018Learning}, STASN~\cite{yang2019face}, GRADIANT~\cite{boulkenafet2017competition} and FAS-TD~\cite{wang2018exploiting}) extracting multi-frame dynamic features, our method needs only frame-level inputs, which is suitable for real-world deployment. It's worth noting that the NAS based backbone for CDCN++ is transferable and generalizes well on all protocols although it is searched on Protocol-1. 

\textbf{Results on SiW.} \quad   Table~\ref{tab:SiW} compares the performance of our method with three state-of-the-art methods Auxiliary~\cite{Liu2018Learning}, STASN~\cite{yang2019face} and FAS-TD~\cite{wang2018exploiting} on SiW dataset. It can be seen from Table~\ref{tab:SiW} that our method performs the best for all three protocols, revealing the excellent generalization capacity of CDC for (1) variations of face pose and expression, (2) variations of different spoof mediums, (3) cross/unknown presentation attack.

 \vspace{-0.4em}
\subsection{Inter Testing}
 \vspace{-0.3em}
\label{sec:Inter}

\newcommand{\tabincell}[2]{\begin{tabular}{@{}#1@{}}#2\end{tabular}}
\begin{table*}
\centering
\caption{AUC (\%) of the model cross-type testing on CASIA-MFSD, Replay-Attack, and MSU-MFSD.}

\scalebox{0.7}{\begin{tabular}{|c|c|c|c|c|c|c|c|c|c|c|}
\hline
\multirow{2}{*}{Method} &\multicolumn{3}{c|}{CASIA-MFSD ~\cite{Zhang2012A}} &\multicolumn{3}{c|}{Replay-Attack ~\cite{ReplayAttack}}&\multicolumn{3}{c|}{MSU-MFSD ~\cite{wen2015face}} &\multirow{2}{*}{Overall} \\
\cline{2-10} &\tabincell{c}{Video} &\tabincell{c}{Cut Photo} &\tabincell{c}{Wrapped Photo} &\tabincell{c}{Video}&\tabincell{c}{Digital Photo}&\tabincell{c}{Printed Photo}&\tabincell{c}{Printed Photo}&\tabincell{c}{HR Video}&\tabincell{c}{Mobile Video} & \\
\hline
OC-SVM$_{RBF}$+BSIF ~\cite{arashloo2017anomaly}
& 70.74 & 60.73 & 95.90 & 84.03 & 88.14 & 73.66 & 64.81 & 87.44 & 74.69 & 78.68$\pm$11.74 \\
\hline
SVM$_{RBF}$+LBP ~\cite{Boulkenafet2017OULU}
& 91.94 & 91.70 & 84.47 & 99.08 & 98.17 & 87.28 & 47.68 & 99.50
 & 97.61 & 88.55$\pm$16.25 \\
\hline
NN+LBP ~\cite{xiong2018unknown}
& 94.16 & 88.39 & 79.85 & 99.75 & 95.17 & 78.86 & 50.57 & 99.93 & 93.54 & 86.69$\pm$16.25 \\
\hline
DTN ~\cite{liu2019deep}
& 90.0 & 97.3 & 97.5 & 99.9 & 99.9 & 99.6 & \textbf{81.6} & 99.9 & 97.5 & 95.9$\pm$6.2 \\
\hline
\textbf{CDCN (Ours)}
& \textbf{98.48} & \textbf{99.90} & \textbf{99.80} & \textbf{100.00} & 99.43 & 99.92 & 70.82 & \textbf{100.00} & \textbf{99.99} & 96.48$\pm$9.64 \\
\hline
\textbf{CDCN++ (Ours)}
& 98.07 & \textbf{99.90} & 99.60 & 99.98 & \textbf{99.89} & \textbf{99.98} & 72.29 & \textbf{100.00} & 99.98 & \textbf{96.63$\pm$9.15} \\
\hline
\end{tabular}
}
\label{tab:cross-type}
\vspace{-1.2em}
\end{table*}

\begin{table}
\centering
\caption{The results of cross-dataset testing between CASIA-MFSD and Replay-Attack. The evaluation metric is HTER(\%). The \textbf{multiple-frame} based methods are shown in the upper half part while \textbf{single-frame} based methods in bottom half part. }
\scalebox{0.81}{\begin{tabular}{|c|c|c|c|c|}
\hline
\multirow{2}{*}{Method} &Train &Test &Train &Test\\
\cline{2-3} \cline{4-5} &\tabincell{c}{CASIA-\\MFSD} &\tabincell{c}{Replay-\\Attack} &\tabincell{c}{Replay-\\Attack} &\tabincell{c}{CASIA-\\MFSD}\\
\hline
Motion-Mag ~\cite{bharadwaj2013computationally}
&\multicolumn{2}{c|}{50.1} &\multicolumn{2}{c|}{47.0} \\
LBP-TOP ~\cite{Pereira2013Can}
&\multicolumn{2}{c|}{49.7} &\multicolumn{2}{c|}{60.6} \\

STASN ~\cite{yang2019face}
&\multicolumn{2}{c|}{31.5} &\multicolumn{2}{c|}{30.9} \\

Auxiliary ~\cite{Liu2018Learning}
&\multicolumn{2}{c|}{27.6} &\multicolumn{2}{c|}{28.4} \\
FAS-TD ~\cite{wang2018exploiting}
&\multicolumn{2}{c|}{17.5} &\multicolumn{2}{c|}{\textbf{24.0}} \\

\hline

LBP ~\cite{boulkenafet2015face}
&\multicolumn{2}{c|}{47.0} &\multicolumn{2}{c|}{39.6} \\
Spectral cubes ~\cite{pinto2015face}
&\multicolumn{2}{c|}{34.4} &\multicolumn{2}{c|}{50.0} \\

Color Texture ~\cite{Boulkenafet2017Face}
&\multicolumn{2}{c|}{30.3} &\multicolumn{2}{c|}{37.7} \\
FaceDs ~\cite{jourabloo2018face}
&\multicolumn{2}{c|}{28.5} &\multicolumn{2}{c|}{41.1} \\

\textbf{CDCN (Ours)}
&\multicolumn{2}{c|}{15.5} &\multicolumn{2}{c|}{32.6} \\
\textbf{CDCN++ (Ours)}
&\multicolumn{2}{c|}{\textbf{6.5}} &\multicolumn{2}{c|}{29.8} \\
\hline
\end{tabular}
}
\label{tab:cross-testing}
\vspace{-0.6em}
\end{table}

To further testify the generalization ability of our model, we conduct cross-type and cross-dataset testing to verify the generalization capacity to unknown presentation attacks and unseen environment, respectively.

\textbf{Cross-type Testing.} \quad
Following the protocol proposed in~\cite{arashloo2017anomaly}, we use CASIA-MFSD~\cite{Zhang2012A}, Replay-Attack~\cite{ReplayAttack} and MSU-MFSD~\cite{wen2015face} to perform intra-dataset cross-type testing between replay and print attacks. As shown in Table~\ref{tab:cross-type}, our proposed CDC based methods achieve the best overall performance (even outperforming the zero-shot learning based method DTN~\cite{liu2019deep}), indicating our consistently good generalization ability among unknown attacks. Moreover, we also conduct the cross-type testing on the latest SiW-M~\cite{liu2019deep} dataset and achieve the best average ACER (12.7\%) and EER (11.9\%) among 13 attacks. The detailed results are shown in \textsl{\textbf{Appendix C}}.

\textbf{Cross-dataset Testing.} \quad   In this experiment, there are two cross-dataset testing protocols. One is that training on the CASIA-MFSD and testing on Replay-Attack, which is named as protocol CR; the second one is exchanging the training dataset and the testing dataset, named protocol RC. As shown in Table~\ref{tab:cross-testing}, our proposed CDCN++ has 6.5\% HTER on protocol CR, outperforming the prior state-of-the-art by a convincing margin of 11\%. For protocol RC, we also outperform state-of-the-art frame-level methods (see bottom half part of Table~\ref{tab:cross-testing} ). The performance might be further boosted via introducing the similar temporal dynamic features in Auxiliary~\cite{Liu2018Learning} and FAS-TD~\cite{wang2018exploiting}.


 \vspace{-0.25em}
\subsection{Analysis and Visualization.}
 \vspace{-0.25em}
 \label{sec:Analysis}

In this subsection, two perspectives are provided to demonstrate the analysis why CDC performs well.
 
\begin{figure}
\includegraphics[width=7.8cm,height=3.3cm]{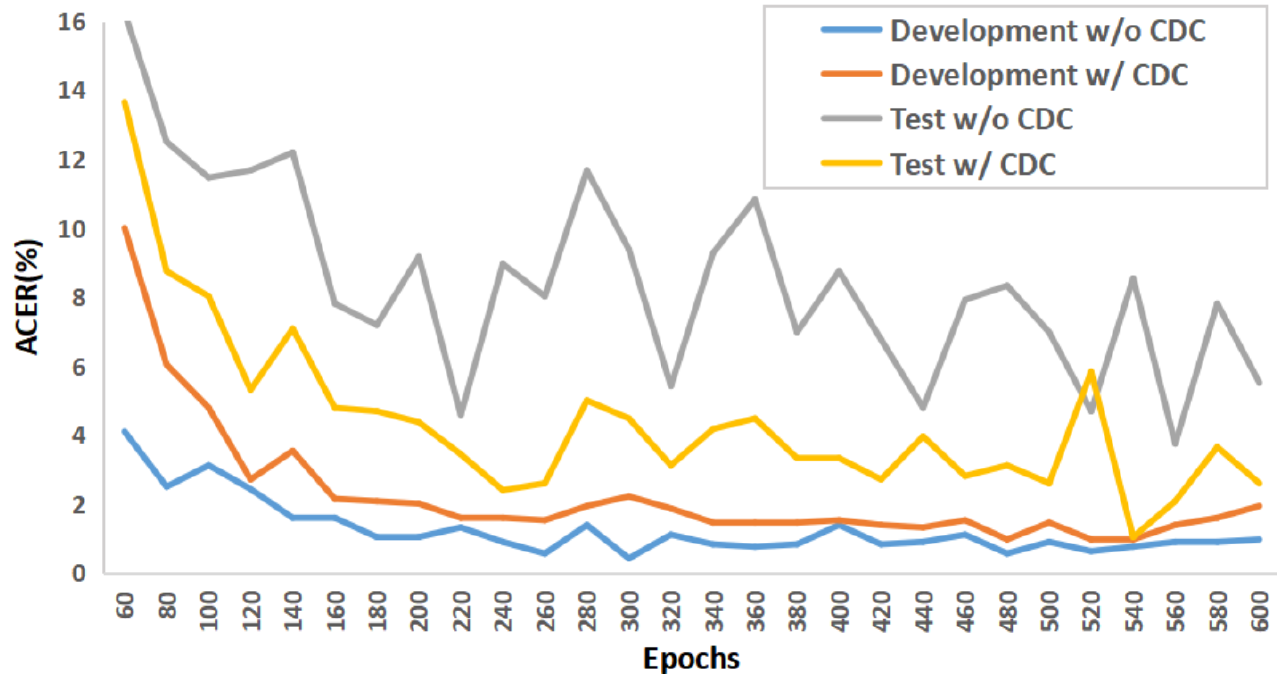}
  \caption{\small{
 The performance of CDCN on development and test set when training on Protocol-1 OULU-NPU. }
  }
 
\label{fig:P1_visual}
\vspace{-1.2em}
\end{figure}

\textbf{Robustness to Domain Shift.}  Protocol-1 of OULU-NPU is used to verify the robustness of CDC when encountering the domain shifting, i.e., huge illumination difference between train/development and test set. Fig.~\ref{fig:P1_visual} shows that the network using vanilla convolution has low ACER on development set (blue curve) while high ACER on test set (gray curve), which indicates vanilla convolution is easily overfitted in seen domain but generalizes poorly when illumination changes. In contrast, the model with CDC is able to achieve more consistent performance on both development (red curve) and test set (yellow curve), indicating the robustness of CDC to domain shifting.

\vspace{-0.2em}

\begin{figure}
\includegraphics[width=8.5cm,height=3.1cm]{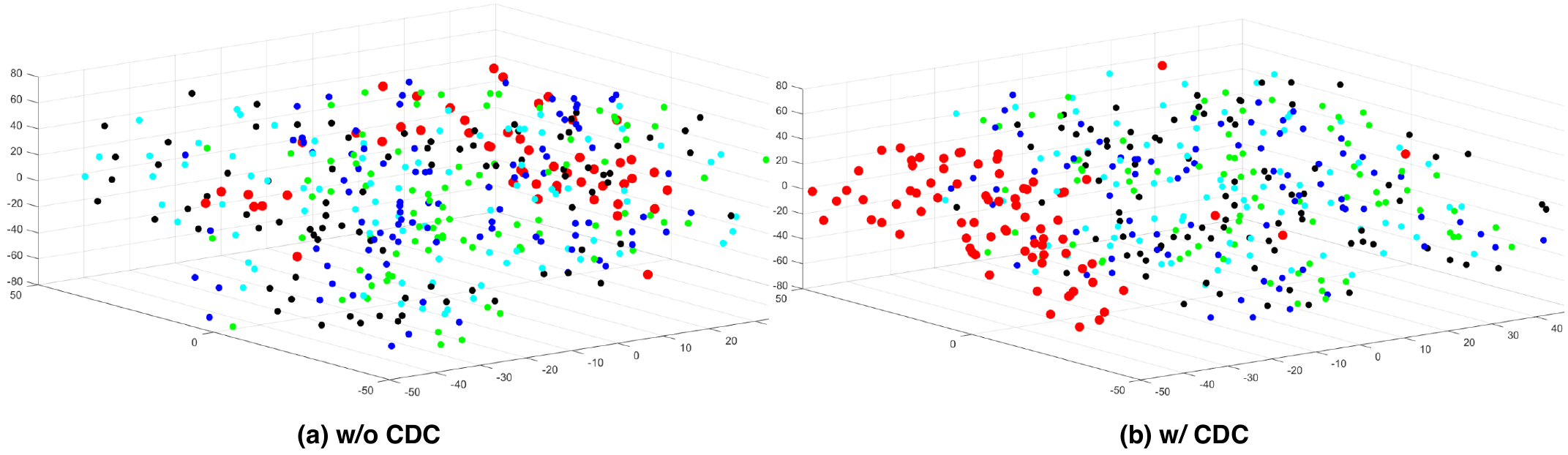}
  \caption{\small{
  3D visualization of feature distribution. (a) Features w/o CDC. (b) Features w/ CDC. Color code used: $red$=live, $green$=printer1, $blue$=printer2, $cyan$=replay1, $black$=replay2.}
  }
\label{fig:distribution}
\vspace{-0.5em}
\end{figure}


\textbf{Features Visualization.} 
Distribution of multi-level fused features for the testing videos on Protocol-1 OULU-NPU is shown in Fig.~\ref{fig:distribution} via t-SNE~\cite{maaten2008visualizing}. It is clear that the features with CDC (Fig.~\ref{fig:distribution}(a)) presents more well-clustered behavior than that with vanilla convolution (Fig.~\ref{fig:distribution}(b)), which demonstrates the discrimination ability of CDC for distinguishing the living faces from spoofing faces. The visualization of the feature maps (w/o or w/ CDC) and attention maps of MAFM can be found in \textsl{\textbf{Appendix D}}.


\vspace{-0.5em}
\section{Conclusions and Future Work}
\vspace{-0.1em}
\label{sec:conc}

In this paper, we propose a novel operator called Central Difference Convolution (CDC) for face anti-spoofing task. Based on CDC, a Central Difference Convolutional Network (CDCN) is designed. We also propose CDCN++, consisting of a searched CDC backbone and Multiscale Attention Fusion Module (MAFM). Extensive experiments are performed to verify the effectiveness of the proposed methods. 
 We note that the study of CDC is still at an early stage. Future directions include: 1) designing context-aware adaptive CDC for each layer/channel;  2) exploring other properties (e.g., domain generalization) and applicability on other vision tasks (e.g., image quality assessment \cite{ling2018role} and FaceForensics).

\section{Acknowledgment}
This work was supported by the Academy of Finland for project MiGA (grant 316765), ICT 2023 project (grant 328115), and Infotech Oulu. As well, the authors wish to acknowledge CSC – IT Center for Science, Finland, for computational resources.

{\small
\bibliographystyle{ieee_fullname}
\bibliography{egbib}
}

\newpage
\newpage

 \section*{\Large Appendix}
\vspace{-0.5em}
\section*{A. Derivation and Code of CDC}
\vspace{-0.1em}
\label{sec:conc}

Here we show the detailed derivation (Eq.(4) in draft) of CDC in Eq.~(\ref{eq:derivation}) and Pytorch code of CDC in Fig.~\ref{fig:code}.

\vspace{-1.0em}
\begin{equation}\scriptsize
\begin{split}
y(p_0)
&=\theta \cdot \underbrace{\sum_{p_n\in \mathcal{R}}w(p_n)\cdot (x(p_0+p_n)-x(p_0))}_{\text{central difference convolution}}\\
&\qquad \qquad \qquad \qquad \qquad \qquad \qquad \qquad+ (1-\theta)\cdot \underbrace{\sum_{p_n\in \mathcal{R}}w(p_n)\cdot x(p_0+p_n)}_{\text{vanilla convolution}}\\
&=\theta \cdot \underbrace{\sum_{p_n\in \mathcal{R}}w(p_n)\cdot x(p_0+p_n)}_{\text{vanilla convolution}}+ \theta\cdot (\underbrace{-\sum_{p_n\in \mathcal{R}}w(p_n)\cdot x(p_0))}_{\text{central difference term}}\\
&\qquad \qquad \qquad \qquad \qquad \qquad \qquad \qquad+ (1-\theta)\cdot \underbrace{\sum_{p_n\in \mathcal{R}}w(p_n)\cdot x(p_0+p_n)}_{\text{vanilla convolution}} \\
&= (\theta+1-\theta)\cdot \underbrace{\sum_{p_n\in \mathcal{R}}w(p_n)\cdot x(p_0+p_n)}_{\text{vanilla convolution}}+\theta\cdot (\underbrace{-x(p_0)\cdot\sum_{p_n\in \mathcal{R}}w(p_n))}_{\text{central difference term}}\\
&= \underbrace{\sum_{p_n\in \mathcal{R}}w(p_n)\cdot x(p_0+p_n)}_{\text{vanilla convolution}}+\theta\cdot (\underbrace{-x(p_0)\cdot\sum_{p_n\in \mathcal{R}}w(p_n))}_{\text{central difference term}}.\\
\end{split}
\label{eq:derivation}
\vspace{-0.6em}
\end{equation}

\vspace{-0.5em}
\begin{lstlisting}
import torch.nn as nn
import torch.nn.functional as F
class CDC (nn.Module):
    def __init__(self, IC, OC, K=3, P=1, theta=0.7):
        # IC, OC: in_channels, out_channels
        # K, P: kernel_size, padding
        # theta: hyperparameter in CDC
        super(CDC, self).__init__() 
        self.vani = nn.Conv2d(IC, OC, kernel_size=K, padding=P)
        self.theta = theta
        
    def forward(self, x):
        # x: input features with shape [N,C,H,W]
        out_vanilla = self.vani(x)
        
        kernel_diff = self.conv.weight.sum(2).sum(2)
        kernel_diff = kernel_diff[:, :, None, None]
        out_CD = F.conv2d(input=x, weight=kernel_diff, padding=0)
        
        return out_vanilla - self.theta * out_CD

\end{lstlisting}
\vspace{-1.5em}
\begin{figure}[h]
\centering
\caption{\small{Python code of CDC based on Pytorch.}}
\label{fig:code}
\vspace{-0.5em}
\end{figure}

\vspace{-0.5em}
\section*{B. Adaptive $\theta$ for CDC}
\vspace{-0.1em}
\label{sec:conc}

Although the best hyperparameter $\theta=0.7$ can be manually measured for face anti-spoofing task, it is still troublesome to find the best-suited $\theta$ when applying Central Difference Convolution (CDC) to other datasets/tasks. Here we treat $\theta$ as the data-driven learnable weights for each layer. A simple implementation is to utilize $Sigmoid(\theta)$ to guarantee the output range within $[0,1]$. 

\begin{figure}[t]
\includegraphics[width=8.8cm,height=3.4cm]{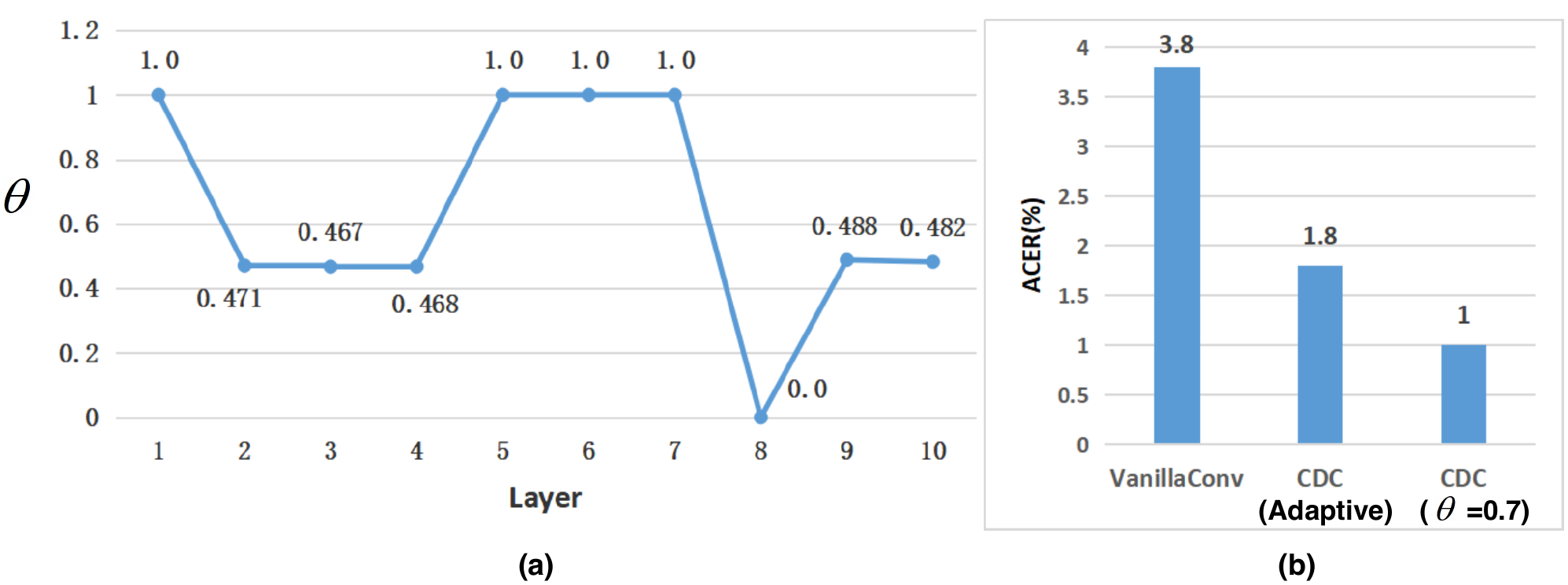}
  \caption{\small{
  Adaptive CDC with learnable $\theta$ for each layer. (a) The learned $\theta$ weights for the first ten layers. (b) Performance comparison on Protocol-1 OULU-NPU.}
  }
 
\label{fig:Adaptive}
\vspace{-0.3em}
\end{figure}

As illustrated in Fig.~\ref{fig:Adaptive}(a), it is interesting to find that the values of learned weights in low (2nd to 4th layer) and high (8th to 10th layer) levels are relatively small while that in mid (5th to 7th layer) level are large. It indicates that the central difference gradient information might be more important for mid level features. In terms of the performance comparison, it can be seen from Fig.~\ref{fig:Adaptive}(b) that adaptive CDC achieves comparable results (1.8\% vs. 1.0\% ACER) with CDC using constant $\theta=0.7$.   


\vspace{1.2em}
\begin{table*}
\centering
\caption{The evaluation and comparison of the cross-type testing on SiW-M~\cite{liu2019deep}.}

\scalebox{0.64}{\begin{tabular}{c|c|c|c|c|c|c|c|c|c|c|c|c|c|c|c}
\hline
\multirow{2}{*}{Method} &\multirow{2}{*}{Metrics(\%)} &\multirow{2}{*}{Replay} &\multirow{2}{*}{Print} &\multicolumn{5}{c|}{Mask Attacks} &\multicolumn{3}{c|}{Makeup Attacks}&\multicolumn{3}{c|}{Partial Attacks} &\multirow{2}{*}{Average} \\
\cline{5-15} &  &  &  & \tabincell{c}{Half} &\tabincell{c}{Silicone} &\tabincell{c}{Trans.} &\tabincell{c}{Paper}&\tabincell{c}{Manne.}&\tabincell{c}{Obfusc.}&\tabincell{c}{Imperson.}&\tabincell{c}{Cosmetic}&\tabincell{c}{Funny Eye} & \tabincell{c}{Paper Glasses} &\tabincell{c}{Partial Paper} & \\
\hline
\hline

\multirow{4}{*}{SVM$_{RBF}$+LBP~\cite{Boulkenafet2017OULU}} & APCER & 19.1 & 15.4 & 40.8 & 20.3 & 70.3 & 0.0 & 4.6 & 96.9 & 35.3 & 11.3 & 53.3 & 58.5 & 0.6 & 32.8$\pm$29.8 \\
\cline{3-16}  & BPCER & 22.1 & 21.5 & 21.9 & 21.4 & 20.7 & 23.1 & 22.9 & 21.7 & 12.5 & 22.2 & 18.4 & 20.0 & 22.9 & 21.0$\pm$2.9 \\
\cline{3-16}  & ACER & 20.6 & 18.4 & 31.3 & 21.4 & 45.5 & 11.6 & 13.8 & 59.3 & 23.9 & 16.7 & 35.9 & 39.2 & 11.7 & 26.9$\pm$14.5 \\
\cline{3-16}  & EER & 20.8 & 18.6 & 36.3  & 21.4 & 37.2 & 7.5 & 14.1 & 51.2 & 19.8 & 16.1 & 34.4 & 33.0 & 7.9 & 24.5$\pm$12.9 \\

\hline
\hline

\multirow{4}{*}{Auxiliary~\cite{Liu2018Learning}} & APCER & 23.7 & 7.3 & 27.7 & 18.2 & 97.8 & 8.3 & 16.2 & 100.0 & 18.0 & 16.3 & 91.8 & 72.2 & 0.4 & 38.3$\pm$37.4 \\
\cline{3-16}  & BPCER & 10.1 & 6.5 & 10.9 & 11.6 & 6.2 & 7.8 & 9.3 & 11.6 & 9.3 & 7.1 & 6.2 & 8.8 & 10.3 & 8.9$\pm$ 2.0 \\
\cline{3-16}  & ACER & 16.8 & 6.9 & 19.3 & 14.9 & 52.1 & 8.0 & 12.8 & 55.8 & 13.7 & \textbf{11.7} & 49.0 & 40.5 & 5.3 & 23.6$\pm$18.5 \\
\cline{3-16}  & EER & 14.0 & 4.3 & 11.6  & 12.4 & 24.6 & 7.8 & 10.0 & 72.3 & 10.1 & \textbf{9.4} & 21.4 & 18.6 & 4.0 & 17.0$\pm$17.7 \\

\hline
\hline

\multirow{4}{*}{DTN~\cite{liu2019deep}} & APCER & 1.0 & 0.0 & 0.7 & 24.5 & 58.6 & 0.5 & 3.8 & 73.2 & 13.2 & 12.4 & 17.0 & 17.0 & 0.2 & 17.1$\pm$23.3 \\
\cline{3-16}  & BPCER & 18.6 & 11.9 & 29.3 & 12.8 & 13.4 & 8.5 & 23.0 & 11.5 & 9.6 & 16.0 & 21.5 & 22.6 & 16.8 & 16.6 $\pm$6.2 \\
\cline{3-16}  & ACER & 9.8 & \textbf{6.0} & 15.0 & 18.7 & 36.0 & 4.5 & 7.7 & 48.1 & 11.4 & 14.2 & \textbf{19.3} & 19.8 & 8.5 & 16.8 $\pm$11.1 \\
\cline{3-16}  & EER & 10.0 & \textbf{2.1} & 14.4 & 18.6 & 26.5 & \textbf{5.7} & 9.6 & 50.2 & 10.1 & 13.2 & \textbf{19.8} & 20.5 & 8.8 & 16.1$\pm$ 12.2 \\

\hline
\hline

\multirow{4}{*}{\textbf{CDCN (Ours)}}& APCER & 8.2 & 6.9 & 8.3 & 7.4 & 20.5 & 5.9 & 5.0 & 43.5 & 1.6 & 14.0 & 24.5 & 18.3 & 1.2 & 12.7$\pm$11.7 \\
\cline{3-16}  & BPCER & 9.3 & 8.5 & 13.9 & 10.9 & 21.0 & 3.1 & 7.0 & 45.0 & 2.3 & 16.2 & 26.4 & 20.9 & 5.4 & 14.6 $\pm$11.7 \\
\cline{3-16}  & ACER & \textbf{8.7} & 7.7 & 11.1 & \textbf{9.1} & 20.7 & 4.5 & 5.9 & 44.2 & 2.0 & 15.1 & 25.4 & 19.6 & 3.3 & 13.6 $\pm$11.7 \\
\cline{3-16}  & EER & \textbf{8.2} & 7.8 & 8.3 & \textbf{7.4} & 20.5 & 5.9 & \textbf{5.0} & 47.8 & 1.6 & 14.0 & 24.5 & 18.3 & 1.1 & 13.1$\pm$ 12.6 \\

\hline
\hline

\multirow{4}{*}{\textbf{CDCN++ (Ours)}} & APCER & 9.2 & 6.0 & 4.2 & 7.4 & 18.2 & 0.0 & 5.0 & 39.1 & 0.0 & 14.0 & 23.3 & 14.3 & 0.0 & 10.8$\pm$11.2 \\
\cline{3-16}  & BPCER & 12.4 & 8.5 & 14.0 & 13.2 & 19.4 & 7.0 & 6.2 & 45.0 & 1.6 & 14.0 & 24.8 & 20.9 & 3.9 & 14.6$\pm$11.4 \\
\cline{3-16}  & ACER & 10.8 & 7.3 & \textbf{9.1} & 10.3 & \textbf{18.8} & \textbf{3.5} & \textbf{5.6} & \textbf{42.1} & \textbf{0.8} & 14.0 & 24.0 & \textbf{17.6} & \textbf{1.9} & \textbf{12.7$\pm$11.2} \\
\cline{3-16}  & EER & 9.2 & 5.6 & \textbf{4.2}  & 11.1 & \textbf{19.3} & 5.9 & \textbf{5.0} & \textbf{43.5} & \textbf{0.0} & 14.0 & 23.3 & \textbf{14.3} & \textbf{0.0} & \textbf{11.9$\pm$11.8} \\

\hline
\hline

\end{tabular}
}
\label{tab:SiW-M}
\vspace{3.2em}
\end{table*}

\begin{figure*}
\centering
\includegraphics[width=17.8cm,height=12.3cm]{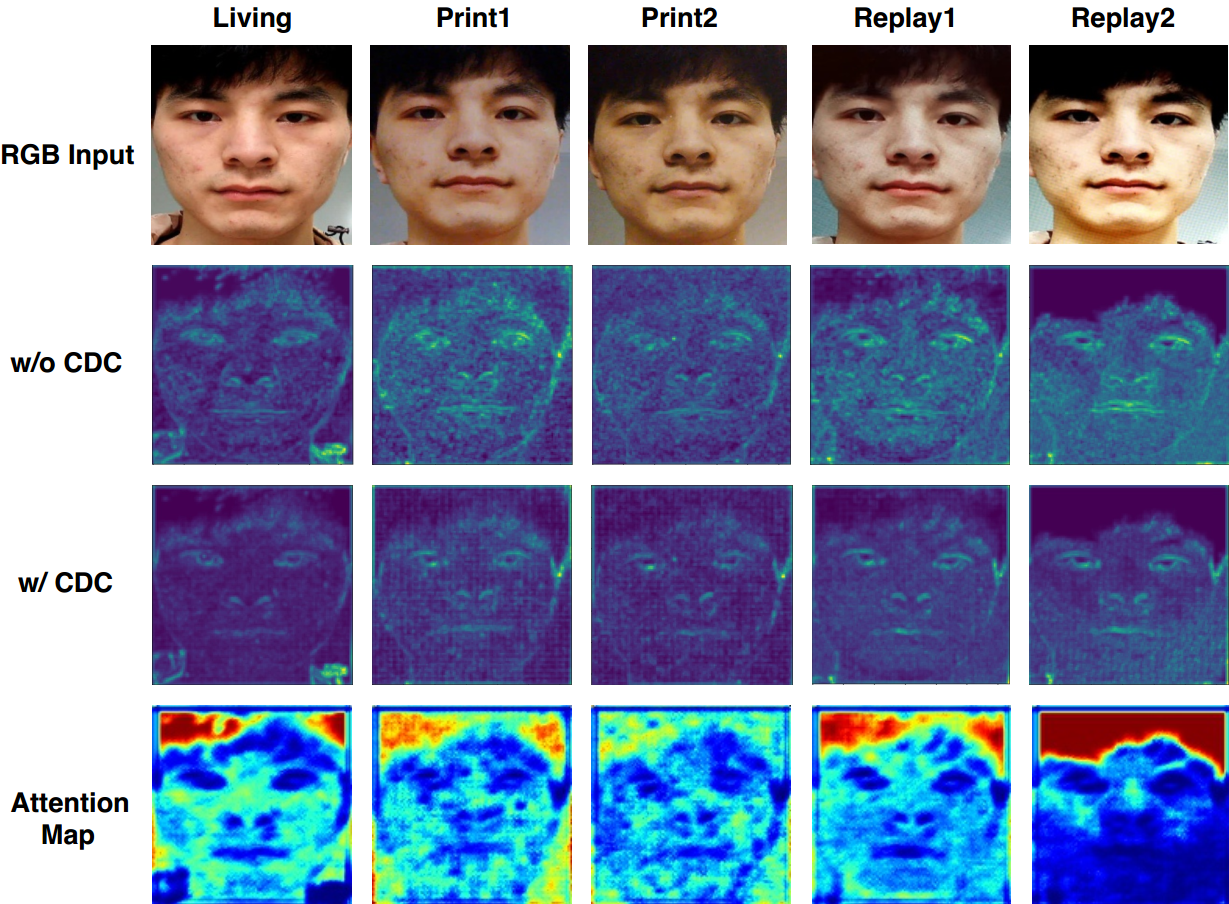}
  \caption{\small{
  Features visualization on living face (the first column) and spoofing faces (four columns to the right). The four rows represent the RGB images, low-level features w/o CDC, w/ CDC and low-level spatial attention maps respectively. Best view when zoom in.}
  }
 
\label{fig:attention}
\end{figure*}

\vspace{-0.5em}
\section*{C. Cross-type Testing on SiW-M}
\vspace{-0.1em}
\label{sec:conc}

Following the same cross-type testing protocol (13 attacks leave-one-out) on SiW-M dataset~\cite{liu2019deep}, we compare our proposed methods with three recent face anti-spoofing methods~\cite{Boulkenafet2017OULU,Liu2018Learning,liu2019deep} to valid the generalization capacity of unseen attacks. As shown in Table ~\ref{tab:SiW-M}, our CDCN++ achieves an overall better ACER and EER, with the improvement of previous state-of-the-art~\cite{liu2019deep} by 24\% and 26\% respectively. Specifically, we detect almost all "Impersonation" and "Partial Paper" attacks (EER=0\%) while the previous methods perform poorly on "Impersonation" attack. It is obvious that 
we reduce the both the EER and ACER of Mask attacks ("HalfMask", "SiliconeMask", "TransparentMask" and "MannequinHead") sharply, which shows our CDC based methods generalize well on 3D nonplanar attacks.

\vspace{-0.5em}
\section*{D. Feature Visualization}
\vspace{-0.1em}
\label{sec:conc}

The low-level features and corresponding spatial attention maps of MAFM are visualized in Fig.~\ref{fig:attention}. It is clear that both the features and attention maps between living and spoofing faces are quite different. 1) For the low-level features (see 2nd and 3rd row in Fig.~\ref{fig:attention}), neural activation from the spoofing faces seems to be more homogeneous between the facial and background regions than that from living faces. It's worth noting that features with CDC are more likely to capture the detailed spoofing patterns (e.g., \textbf{lattice artifacts} in "Print1" and \textbf{reflection artifacts} in "Replay2"). 2) For the spatial attention maps of MAFM (see 4th row in Fig.~\ref{fig:attention}), all the regions of hair, face and background have the relatively strong activation for the living faces while the facial regions contribute weakly for the spoofing faces. 


\end{document}